\documentclass[runningheads]{llncs}
\usepackage{graphicx}
\usepackage{amsmath}
\usepackage{bm}
\usepackage[noabbrev]{cleveref}
\usepackage{subfig}
\usepackage{amssymb}
\usepackage{caption}
\usepackage{xcolor}
\usepackage[normalem]{ulem}

\newcommand{\Win}{\bm W_\mathrm{in}}
\newcommand{\Wout}{\bm W_\mathrm{out}}
\newcommand{\W}{\bm W}

\begin{document}
\title{Learning ergodic averages in chaotic systems\thanks{F.H. acknowledges the support from Fundação para a Ciência e Tecnologia under the Research Studentship no. SFRH/BD/134617/2017. L.M. acknowledges the support of the Technical University of Munich -- Institute for Advanced Study, funded by the German Excellence Initiative and the European Union Seventh Framework Programme under grant agreement no. 291763.}}
%
%
\author{Francisco Huhn\inst{1} \and Luca Magri\inst{1,2}}
\authorrunning{F. Huhn et al.}
%
\institute{Department of Engineering, University of Cambridge, United Kingdom \and
Technical University of Munich, Germany (visiting)}
\maketitle
\begin{abstract}
We propose a physics-informed machine learning method to predict the time average of a chaotic attractor. The method is based on the hybrid echo state network (hESN). We assume that the system is ergodic, so the time average is equal to the ergodic average. Compared to conventional echo state networks (ESN) (purely data-driven), the hESN uses additional information from an incomplete, or imperfect, physical model. We evaluate the performance of the hESN and compare it to that of an ESN. This approach is demonstrated on a chaotic time-delayed thermoacoustic system, where the inclusion of a physical model significantly improves the accuracy of the prediction, reducing the relative error from 48\% to 1\%. This improvement is obtained at the low extra cost of solving a small number of ordinary differential equations that contain physical information. 
This framework shows the potential of using machine learning techniques combined with prior physical knowledge to improve the prediction of time-averaged quantities in chaotic systems.
\keywords{Echo State Networks \and Hybrid Echo State Networks \and Physics-Informed Echo State Networks \and Chaotic Dynamical Systems}
\end{abstract}
\section{Introduction}

In the past decade, there has been a proliferation of machine learning techniques applied in various fields, from spam filtering \cite{Guzella2009} to self-driving cars \cite{Bojarski2016}, including the more recent physical applications in fluid dynamics \cite{Duraisamy2019,Brunton2020}. However, a major hurdle in applying machine learning to complex physical systems, such as those in fluid dynamics, is the high cost of generating data for training \cite{Duraisamy2019}. Nevertheless, this can be mitigated by leveraging prior knowledge (e.g. physical laws). Physical knowledge can compensate for the small amount of training data.
These approaches, called physics-informed machine learning, have been applied to various problems in fluid dynamics \cite{Duraisamy2019,Brunton2020}. For example, \cite{Pathak2018,Doan2019} improve the predictability horizon of echo state networks by leveraging physical knowledge, which is enforced as a {\it hard} constraint in~\cite{Doan2019}, without needing more data or neurons.
In this study, we use a hybrid echo state network (hESN) \cite{Pathak2018}, originally proposed to time-accurately forecast the evolution of chaotic dynamical systems, to predict the long-term time averaged quantities, i.e., the ergodic averages. This is motivated by recent research in optimization of chaotic multi-physics fluid dynamics problems with applications to thermoacoustic instabilities~\cite{Huhn2020}. The hESN is based on reservoir computing \cite{Lukosevicius2009}, in particular, conventional Echo State Networks (ESNs). ESNs have shown to predict nonlinear and chaotic dynamics more accurately and for a longer time horizon than other deep learning algorithms \cite{Lukosevicius2009}. However, we stress that the present study is not focused on the accurate prediction of the time evolution of the system, but rather of its ergodic averages, which are obtained by the time averaging of a long time series (we implicitly assume that the system is ergodic, thus, the infinite time average is equal to the ergodic average~\cite{Birkhoff1931}.).
Here, the physical system under study is a prototypical time-delayed thermoacoustic system, whose chaotic dynamics have been analyzed and optimized in \cite{Huhn2020}.

\section{Echo State Networks}
\label{sec:echo_state_networks}

The ESN approach presented in \cite{Lukosevicius2012} is used here. The ESN is given an input signal $\bm u(n) \in \mathbb{R}^{N_u}$, from which it produces a prediction signal $\hat{\bm y}(n) \in \mathbb{R}^{N_y}$ that should match the target signal $\bm y(n) \in \mathbb{R}^{N_y}$, where $n$ is the discrete time index. The ESN is composed of a reservoir, which can be represented as a directed weighted graph with $N_x$ nodes, called neurons, whose state at time $n$ is given by the vector $\bm x(n)  \in \mathbb{R}^{N_x}$. The reservoir is coupled to the input via an input-to-reservoir matrix, $\Win$, such that its state evolves according to
\begin{equation}
    \label{eq:reservoir}
    \bm x(n) = \tanh(\Win \bm u(n) + \W \bm x(n-1)),
\end{equation}
where $\W \in \mathbb{R}^{N_x} \times \mathbb{R}^{N_x}$ is the weighted adjacency matrix of the reservoir, i.e. $W_{ij}$ is the weight of the edge from node $j$ to node $i$, and the hyperbolic tangent is the activation function. Finally, the prediction is produced by a linear combination of the states of the neurons
\begin{equation}
    \hat{\bm y}(n) = \Wout \bm x(n),
\end{equation}
where $\Wout \in \mathbb{R}^{N_y} \times \mathbb{R}^{N_x}$. In this work, we are interested in dynamical system prediction. Thus, the target at time step $n$ is the input at time step $n+1$, i.e. $\bm y(n) = \bm u(n+1)$ \cite{Pathak2018}.
We wish to learn ergodic averages, given by
\begin{equation}
    \langle \mathcal{J} \rangle = \lim_{T \rightarrow \infty} \frac{1}{T} \int_0^T \mathcal{J}(\bm u(t)) \, dt,
\end{equation}
where $\mathcal{J}$ is a cost functional, of a dynamical system governed by
\begin{equation}
    \label{eq:FOM}
    \dot{\bm u} = \bm F(\bm u),
\end{equation}
where $\bm u \in \mathbb{R}^{N_u}$ is the state vector and $\bm F$ is a nonlinear operator. The training data is obtained via numerical integration of \cref{eq:FOM}, resulting in the time series $\{\bm u(1), \dots, \bm u(N_t)\}$, where the different samples are taken at equally spaced time intervals $\Delta t$, and $N_t$ is the length of the training data set. In the conventional ESN approach, $\Win$ and $\W$ are generated once and fixed. Then, $\W$ is re-scaled to have the desired spectral radius, $\rho$, to ensure that the network satisfies the Echo State Property \cite{Jaeger2004}. Only $\Wout$ is trained to minimize the mean-squared-error
\begin{equation}
    \label{eq:MSE}
    E_d = \frac{1}{N_y} \sum_{i=1}^{N_y} \frac{1}{N_t} \sum_{n=1}^{N_t} (\hat{y}_i(n) - y_i(n))^2.
\end{equation}
To avoid overfitting, we use ridge regularization, so the optimization problem is
\begin{equation}
    \underset{\Wout}{\text{min}} \; E_d + \gamma \vert\vert \Wout \vert\vert^2,
\end{equation}
where $\gamma$ is the regularization factor. Because the prediction $\hat{\bm y}(n)$ is a linear combination of the reservoir state $\bm x(n)$, the optimal $\Wout$ can be explicitly obtained with
\begin{equation}
    \Wout = \bm Y \bm X^T (\bm X \bm X^T + \gamma \bm I)^{-1},
\end{equation}
where $\bm I$ is the identity matrix and $\bm Y$ and $\bm X$ are the column-concatenation of the various time instants of the output data, $\bm y$, and corresponding reservoir states, $\bm x$, respectively.
After the optimal $\Wout$ is found, the ESN can be used to predict the time evolution of the system. This is done by looping back its output to its input, i.e. $\bm u(n) = \hat{\bm y}(n-1) = \Wout \bm x(n-1)$, which, on substitution into \cref{eq:reservoir}, results in
\begin{equation}
    \bm x(n) = \tanh( \widetilde{\W} \bm x(n-1) ), \label{eq:esn_sym}
\end{equation}
with $\widetilde{\bm W}=\bm W + \Win \Wout$.
Interestingly, \cref{eq:esn_sym} shows that if the reservoir follows an evolution of states $\bm x(1), \dots, \bm x(N_p)$, where $N_p$ is the number of prediction steps, then $-\bm x(1), \dots, -\bm x(N_p)$ is also possible, because flipping the sign of $\bm x$ in \cref{eq:esn_sym} results in the same equation. This implies that either the attractor of the ESN (if any) is symmetric, i.e. if some $\bm x$ is in the ESN's attractor, then so is $-\bm x$; or the ESN has a co-existing symmetric attractor. While this seemed not to have been an issue in short-term prediction, such as in \cite{Pathak2018,Doan2019}, it does pose a problem in the long-term prediction of statistical quantities. This is because the ESN, in its present form, \textit{can not} generate non-symmetric attractors. This symmetry needs to be broken to work with a general non-symmetric dynamical system. This can be done by including biases \cite{Lukosevicius2012}. However, the addition of a bias can make the reservoir prone to saturation (results not shown), i.e. $x_i \rightarrow \pm 1$, and thus care needs to be taken in the choice of hyperparameters.
In this paper, we break the symmetry by exploiting prior knowledge on the physics of the problem under investigation with a hybrid ESN.

\section{Physics-informed and hybrid Echo State Network}

\begin{figure}[tb]
    \centering
    \includegraphics[width=0.9\textwidth]{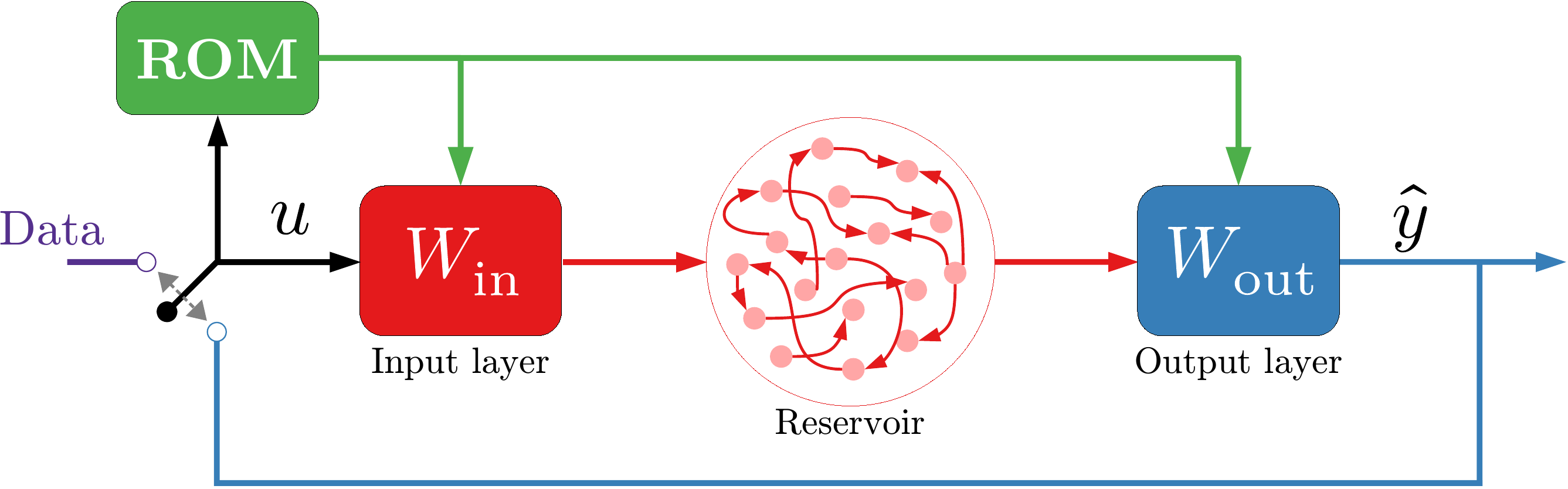}
    \caption{Schematic of the hybrid echo state network. In training mode, the input of the network is the training data (switch is horizontal). In prediction mode, the input of the network is its output from the previous time step (switch is vertical).}
    \label{fig:hybrid}
\end{figure}

The ESN's performance can be increased by incorporating physical knowledge during training \cite{Doan2019} or during trainng and prediction \cite{Pathak2018}. This physical knowledge is usually present in the form of a reduced-order model (ROM) that can generate (imperfect) predictions. The authors of \cite{Doan2019} introduced a physics-informed ESN (PI-ESN), which constrains the physics as a \textit{hard} constraint with a physics loss term.
The prediction is consistent with the physics, but the training requires nonlinear optimization.
The authors of \cite{Pathak2018} introduced a hybrid echo state network (hESN), which incorporates incomplete physical knowledge by feeding the prediction of the physical model into the reservoir and into the output. This requires ridge regression.
Here, we use a hESN (Figure~\ref{fig:hybrid}) because we are not interested in constraining the physics as a hard constraint for an accurate short-term prediction~\cite{Doan2019}. In the hESN, similarly to the conventional ESN, the input is fed to the reservoir via the input layer $\Win$, but also to a physical model, which is usually a set of ordinary differential equations that approximately describe the system that is to be predicted. In this work, that model is a reduced-order model (ROM) of the full system. The output of the ROM is then fed to the reservoir via the input layer and into the output of the network via the output layer. 

\section{Learning the ergodic average of an energy}
\label{sec:results}

\begin{figure}[tb]
    \centering
    \includegraphics[width=0.47\textwidth]{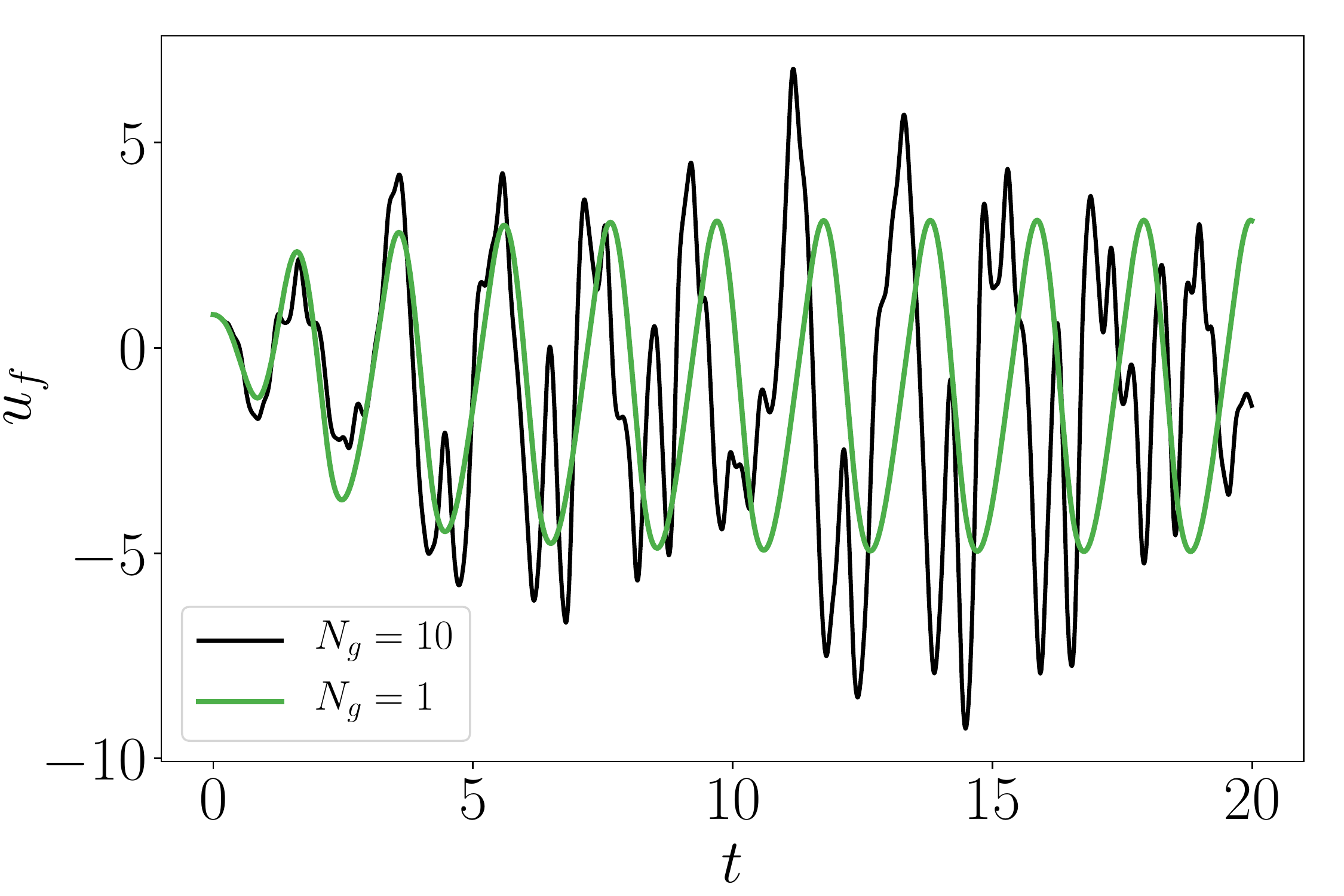}
    \caption{Acoustic velocity at the flame location.}
    \label{fig:uf}
\end{figure}

We use a prototypical time-delayed thermoacoustic system composed of a longitudinal acoustic cavity and a heat source modelled with a nonlinear time-delayed model~\cite{Nair2015,Traverso2019,Huhn2020}, which has been used to optimize ergodic averages in \cite{Huhn2020} with a dynamical systems approach.
The non-dimensional governing equations are
\begin{equation}
	\partial_t u + \partial_x p = 0, \quad \partial_t p + \partial_x u + \zeta p - \dot{q}\delta(x - x_f)= 0, \label{eq:rijke}
\end{equation}
where $u$, $p$, $\zeta$ and $\dot q$ are the non-dimensionalized acoustic velocity, pressure, damping and heat-release rate, respectively. $\delta$ is the Dirac delta.
These equations are discretized by using $N_g$ Galerkin modes
\begin{equation}
    u(x,t) = \sum\nolimits_{j=1}^{N_g} \eta_j(t)\cos(j \pi x), \quad p(x,t) = -\sum\nolimits_{j=1}^{N_g} \mu_j(t) \sin(j \pi x), \label{eq:galerkin_decomp}
\end{equation}
which results in a system of $2N_g$ oscillators, which are nonlinearly coupled through the heat released by the heat source
\begin{equation}
	\dot{\eta}_j - j \pi \mu_j = 0, \quad \dot{\mu}_j + j \pi \eta_j + \zeta_j \mu_j + 2 \dot{q} \sin(j \pi x_f) = 0 \label{eq:rijke_gal},
\end{equation}
where $x_f=0.2$ is the heat source location and $\zeta_j = 0.1 j + 0.06 j^{1/2}$ is the modal damping \cite{Huhn2020}. The heat release rate, $\dot q$, is given by the modified King's law \cite{Huhn2020}, $\dot{q}(t) = \beta [ \left(1+u(x_f, t-\tau)\right)^{1/2} - 1 ]$, where $\beta$ and $\tau$ are the heat release intensity parameter and the time delay, respectively. 
With the nomenclature of \cref{sec:echo_state_networks}, $\bm y(n) = (\eta_1; \dots; \eta_{N_g}; \mu_1 ; \dots; \mu_{N_g})$.
Using 10 Galerkin modes ($N_g=10$), $\beta=7.0$ and $\tau=0.2$ results in a chaotic motion (\cref{fig:uf}), with the leading Lyapunov exponent being $\lambda_1 \approx 0.12$ \cite{Huhn2020}. (The leading Lyapunov exponent measures the rate of (exponential) separation of two close initial conditions, i.e. an initial separation $||\bm{\delta u}_0||$ grows asymptotically like $||\bm{\delta u}_0|| e^{\lambda_1 t}$.) However, for the same choice of parameter values, the solution with $N_g=1$ is a limit cycle (i.e. a periodic solution).

The echo state network is trained on data generated with $N_g=10$, while the physical knowledge (ROM in \cref{fig:hybrid}) is generated with $N_g=1$ only.
We wish to predict the time average of the instantaneous acoustic energy,
\begin{equation}
E_{ac}(t)=\int_0^1 \frac{1}{2}(u^2 + p^2) \, dx,
\end{equation}
which is a relevant metric in the optimization of thermoacoustic systems \cite{Huhn2020}. 
The reservoir is composed of 100 units, a modest size, half of which receive their input from $\bm u$, while the other half receives it from the output of the ROM, $ \hat{\bm y}_\text{ROM}$.
The entries of $\Win$ are randomly generated from the uniform distribution $\text{unif}(-\sigma_\mathrm{in}, \sigma_\mathrm{in})$, where $\sigma_\mathrm{in} = 0.2$. The matrix $\W$ is highly sparse, with only 3\% of non-zero entries from the uniform distribution  $\text{unif}(-1, 1)$. Finally, $\W$ is scaled such that its spectral radius, $\rho$, is 0.1 and 0.3 for the ESN and the hESN, respectively. The time step is $\Delta t = 0.01$. The network is trained for $N_t = 5000$ units, which corresponds to 6 Lyapunov times, i.e. $6\lambda_1^{-1}$. The data is generated by integrating \cref{eq:rijke_gal} in time with $N_g=10$, resulting in $N_u = N_y = 20$. In the hESN, the ROM is obtained by integrating the same equations, but with $N_g=1$ (one Galerkin mode only) unless otherwise stated. Ridge regression is performed with $\gamma=10^{-7}$. The values of the hyperparameters are taken from the literature \cite{Pathak2018,Doan2019} and a grid search, which is, while not the most efficient, is well suited when there are few hyperparameters, such as this work's ESN architecture.

\begin{figure}[htb]
    \centering
    \subfloat[Absolute error of $(\eta_1; \mu_1)$ prediction.]{
        \label{fig:inst_pred}
        \includegraphics[width=0.47\textwidth]{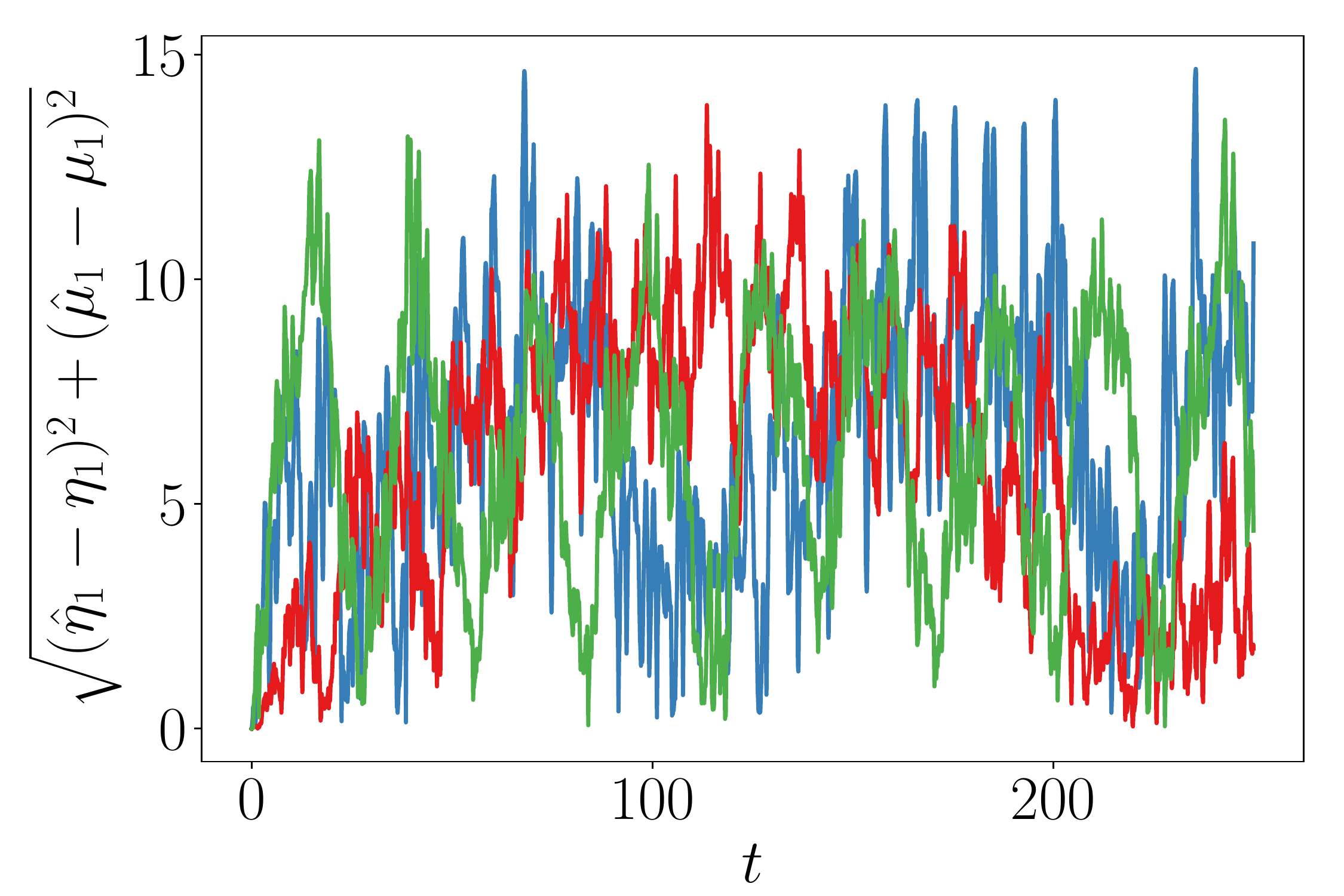}
    }
    \subfloat[Relative error of $\langle E_{ac} \rangle$ prediction.]{
        \label{fig:stats_pred}
        \includegraphics[width=0.47\textwidth]{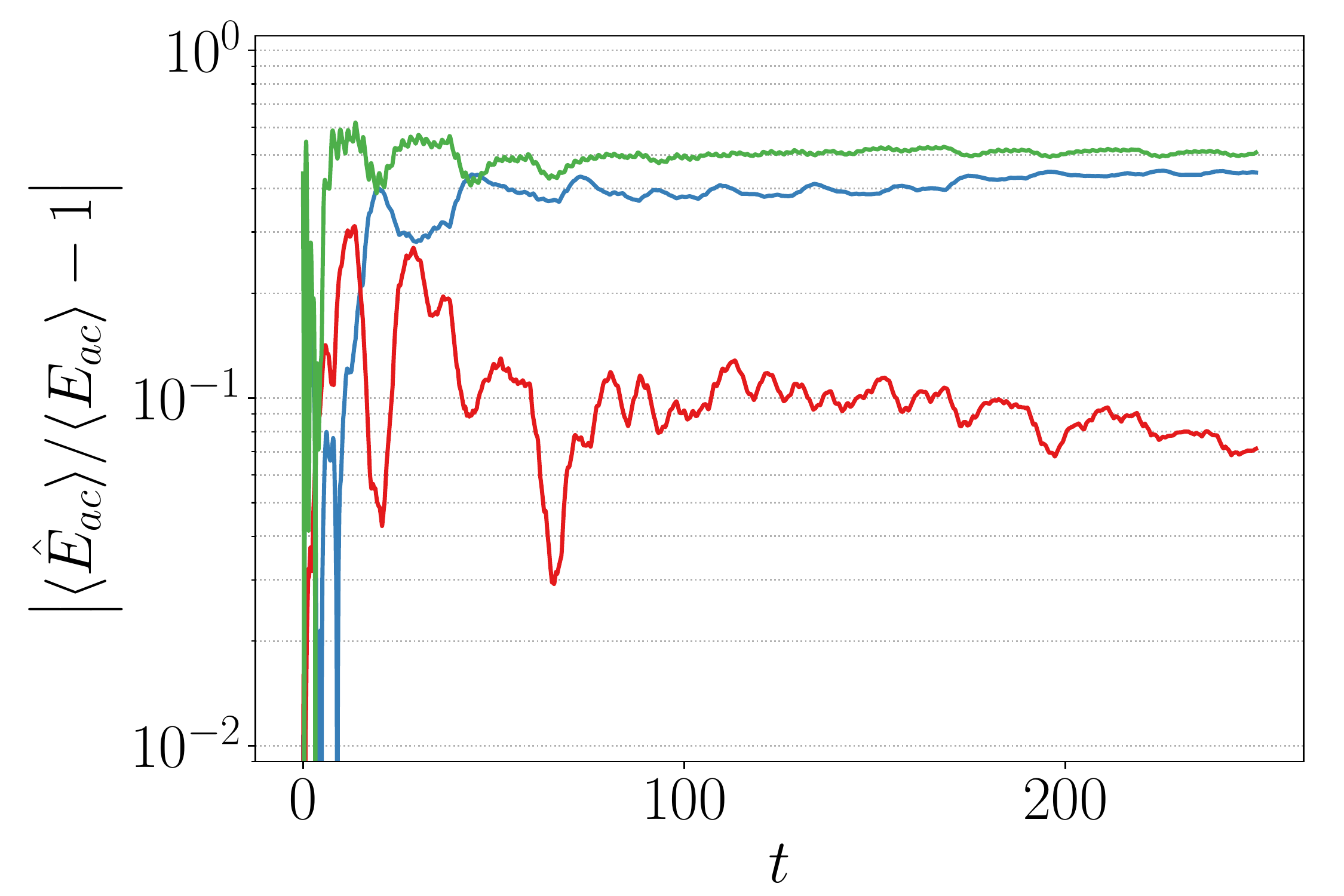}
    }
    
    \subfloat[Relative error vs number of Galerkin modes of the ROM.]{
        \label{fig:ng_study}
        \includegraphics[width=0.47\textwidth]{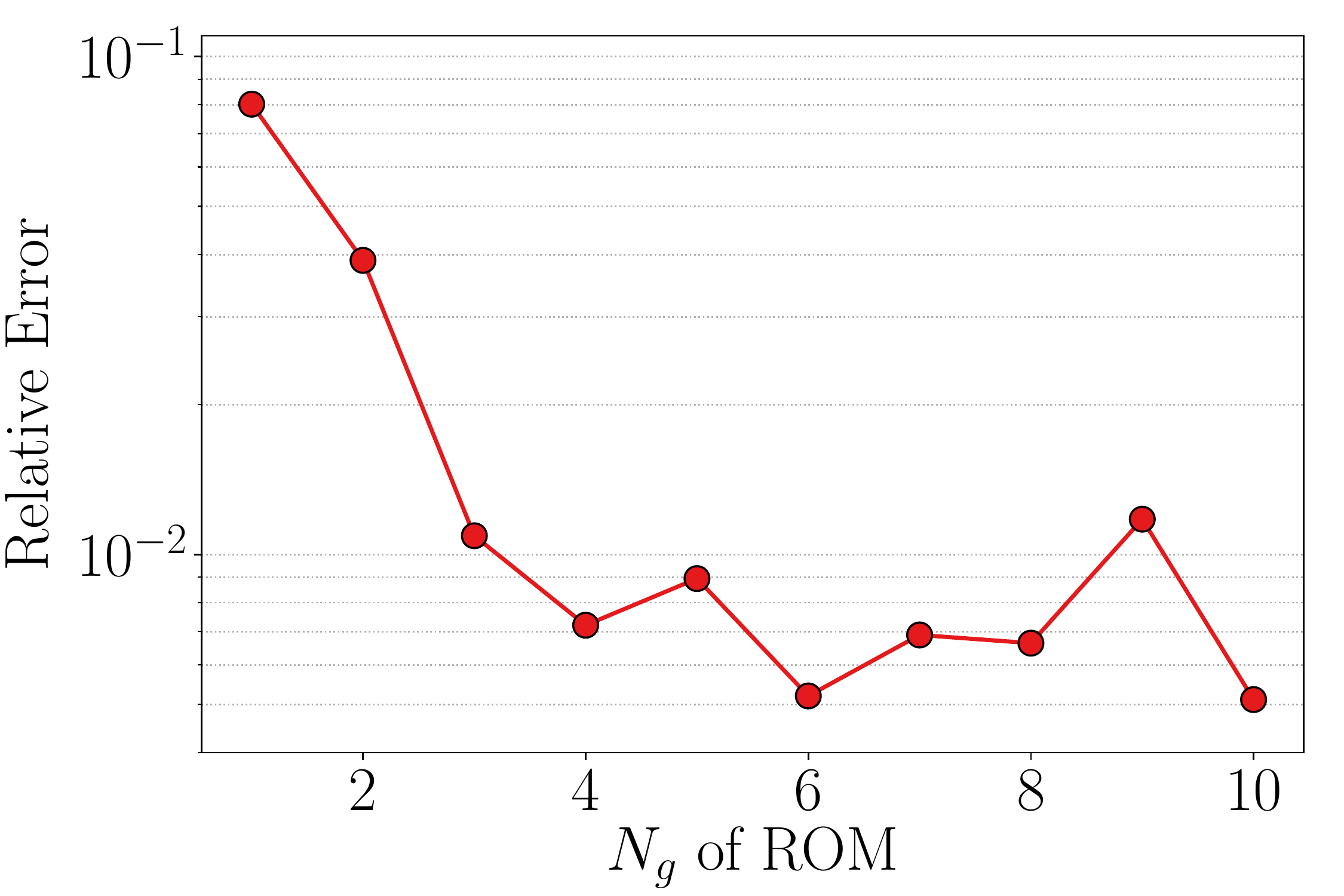}
    }
    \caption{Errors on the prediction from ESN (blue), hESN (red) and ROM (green).}
    \label{fig:preds}
\end{figure}

On the one hand, \cref{fig:inst_pred} shows the instantaneous error of the first modes of the acoustic velocity and pressure $(\eta_1; \mu_1)$ for the ESN, hESN and ROM. None of these can accurately predict the instantaneous state of the system. On the other hand, \cref{fig:stats_pred} shows the error of the prediction of the average acoustic energy. Once again, the ROM alone does a poor job at predicting the statistics of the system, with an error of 50\%. This should not come at a surprise since, as discussed previously, the ROM does not even produce a chaotic solution. The ESN, trained on data only, performs marginally better, with an error of 48\%. In contrast, the hESN predicts the time-averaged acoustic energy satisfactorily, with an error of about 7\%. This is remarkable, since both the ESN and the ROM do a poor job at predicting the average acoustic energy. However, when the ESN is combined with prior knowledge from the ROM, the prediction becomes significantly better. Moreover, while the hESN's error still decreases at the end of the prediction period, $t=250$, which is 5 times the training data time, the ESN and the ROM stabilize much earlier, at a time similar to that of the training data. This result shows that complementing the ESN with a cheap physical model (only 10\% the number of degrees of freedom of the full system) can greatly improve the accuracy of the predictions, with no need for more data or neurons.
\Cref{fig:ng_study} shows the relative error as a function of the number of Galerkin modes in the ROM, which is a proxy for the quality of the model. For each $N_g$, we take the median of $16$ reservoir realizations. As expected, as the quality of the model increases, so does the quality of the prediction. This effect is most noticeable from $N_g=1$ to 4, with the curve presenting diminishing returns. The downside of increasing $N_g$ is obviously the increase in computational cost. At $N_g=10$, the original system is recovered. However, the error does not tend exactly to 0 because $\Wout$ can not combine the ROM's output only (i.e. 0 entries for reservoir nodes) due to: i) the regularization factor in ridge regression that penalizes large entries; ii) numerical error. This graph further strengthens the point previously made that cheap physical models can greatly improve the prediction of physical systems with data techniques.
%

\begin{figure}[htb]
    \centering
    \subfloat[($\beta=7.0$, $\tau=0.2$): $\sigma_\mathrm{in}=0.2$]{
        \label{fig:phase_chaotic}
        \includegraphics[width=0.47\textwidth]{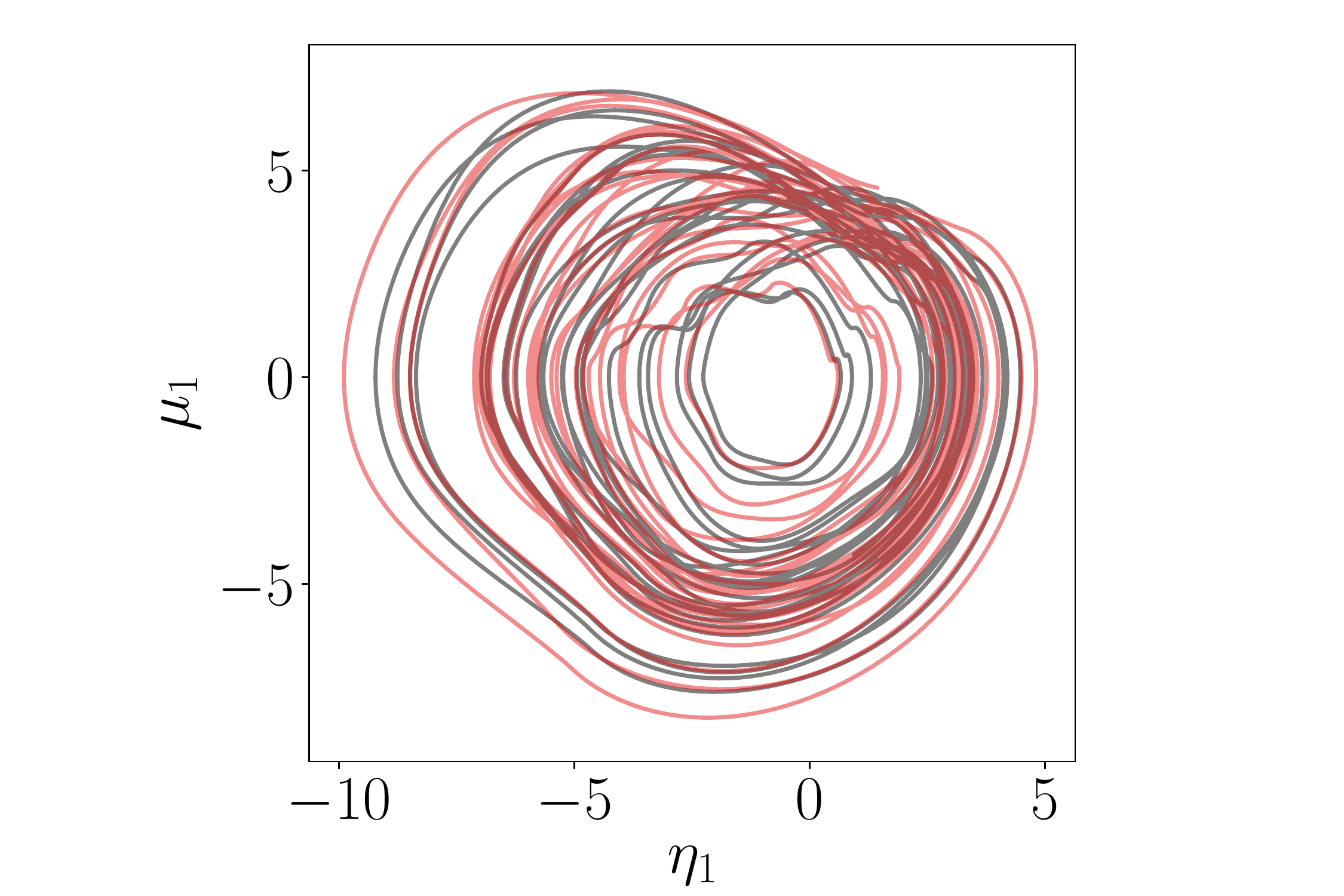}
    }
    \subfloat[($\beta=6.0$, $\tau=0.3$): $\sigma_\mathrm{in}=0.03$]{
        \label{fig:phase_periodic}
        \includegraphics[width=0.47\textwidth]{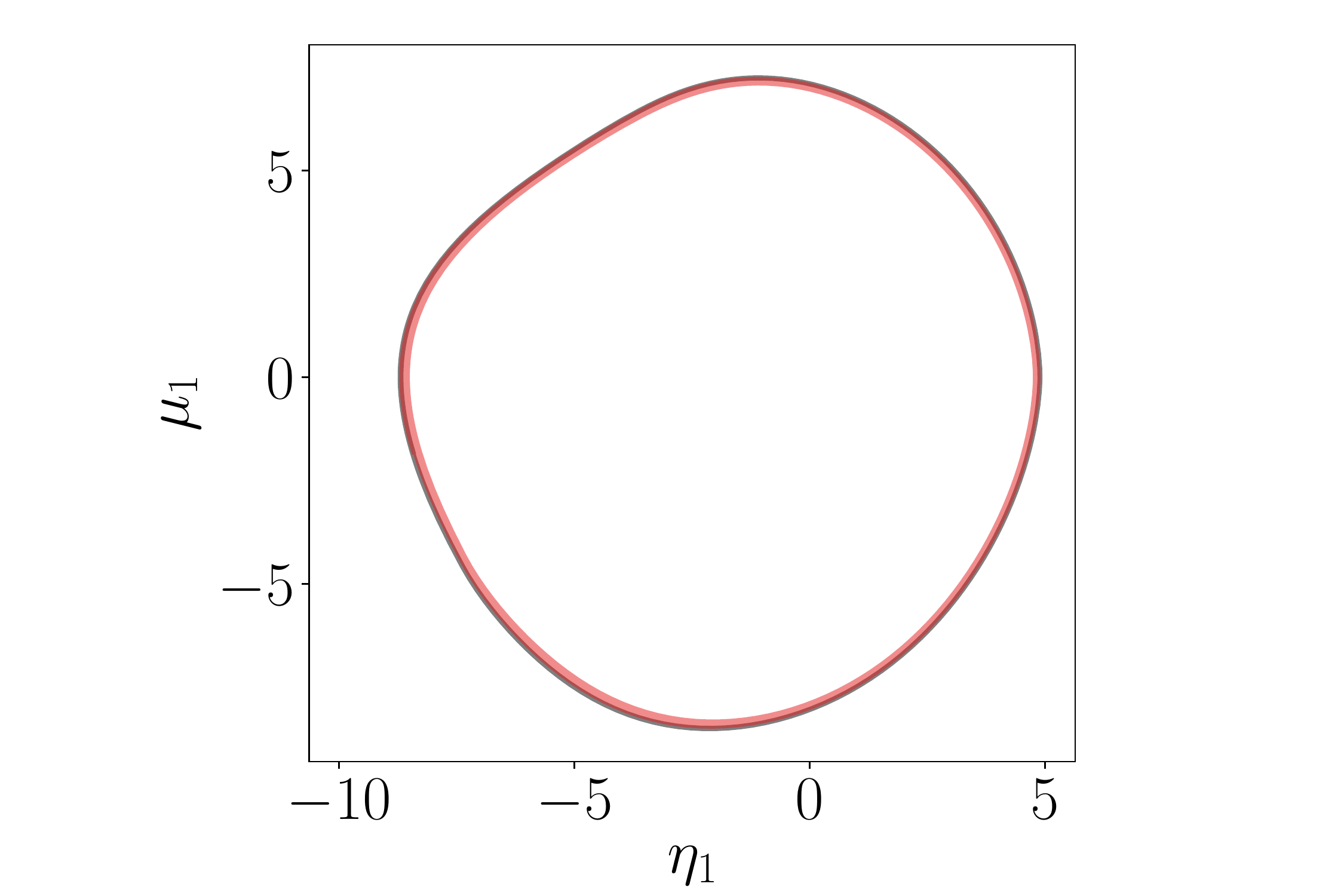}
    }
    \caption{Phase plot of system (black) and hESN (red) for two sets of $(\beta,\tau)$. (a) chaotic solution; (b) periodic solution.}
    \label{fig:phase}
\end{figure}

We stress that the optimal values of hyperparameters for a certain set of physical parameters, e.g. $(\beta_1, \tau_1)$, might not be optimal for a different set of physical parameters $(\beta_2, \tau_2)$. This should not be surprising, since different physical parameters will result in different attractors.
For example, \cref{fig:phase} shows that changing the physical parameters from $(\beta=7.0, \tau=0.2)$ to $(\beta=6.0, \tau=0.3)$ results in a change of type of attractor from chaotic to limit cycle. For the hESN to predict the limit cycle, the value of $\sigma_\mathrm{in}$ must change from $0.2$ to $0.03$
Thus, if the hESN (or any deep learning technique in general) is to be used to predict the dynamics of various physical configurations (e.g. the generation of a bifurcation diagram), then it should be coupled with a robust method for the automatic selection of optimal hyperparameters~\cite{Bengio2012}, with a promising candidate being Bayesian optimization~\cite{Yperman2016,Maat2019}.

\section{Conclusion and future directions}
\label{sec:conclusion}

We propose the use of echo state networks informed with incomplete prior physical knowledge for the prediction of time averaged cost functionals in chaotic dynamical systems. We apply this to chaotic acoustic oscillations, which is relevant to aeronautical propulsion. The inclusion of physical knowledge comes at a low cost and significantly improves the performance of conventional echo state networks from a 48\% error to 1\%, without requiring additional data or neurons. This improvement is obtained at the low extra cost of solving a small number of ordinary differential equations that contain physical information. 
The ability of the proposed ESN can be exploited in the optimization of chaotic systems by accelerating computationally expensive shadowing methods \cite{Ni2017}.
For future work, (i) the performance of the hybrid echo state network should be compared against those of other physics-informed machine learning techniques; (ii) robust methods for hyperparameters' search should be coupled for a ``hands-off'' autonomous tool; and (iii) this technique is currently being applied to larger scale problems.
In summary, the proposed framework is able to learn the ergodic average of a fluid dynamics system, which opens up new possibilities for the optimization of highly unsteady problems.

%
%
%
\bibliographystyle{abbrv}
\bibliography{biblio.bib}

\end{document}